\def\year{2020}\relax
\documentclass[letterpaper]{article} 
\usepackage{aaai20}  
\usepackage{times}  
\usepackage{helvet} 
\usepackage{courier}  
\usepackage[hyphens]{url}  
\usepackage{graphicx} 
\usepackage{graphicx}  
\usepackage{algorithm}
\usepackage[noend]{algpseudocode}
\usepackage{booktabs}       

\usepackage{multirow}
\usepackage{graphicx}
\usepackage{xcolor}
\usepackage{amsthm}
\usepackage{amsmath,amssymb}
\DeclareMathOperator*{\argmax}{argmax}

\newtheorem{theorem}{Theorem}

\newtheorem{lemma}{Lemma}

\title{Improving Policies via Search in Cooperative Partially Observable Games}

\author{%
  Adam Lerer \\
  Facebook AI Research\\
  \texttt{alerer@fb.com} \\
  \And
  Hengyuan Hu \\
  Facebook AI Research\\
  \texttt{hengyuan@fb.com} \\
  \And
  Jakob Foerster \\
  Facebook AI Research\\
  \texttt{jnf@fb.com} \\
  \And
  Noam Brown \\
  Facebook AI Research\\
  \texttt{noambrown@fb.com} \\
}

\begin{document}
\maketitle
\begin{abstract}
\vspace{-1mm}
Recent superhuman results in games have largely been achieved in a variety of zero-sum settings, such as Go and Poker, in which agents need to compete against others. 
However, just like humans, real-world AI systems have to coordinate and communicate with other agents in cooperative partially observable environments as well.
These settings commonly require participants to both interpret the actions of others and to act in a way that is informative when being interpreted. Those abilities are typically summarized as \emph{theory of mind} and are seen as crucial for social interactions. 
In this paper we propose two different search techniques that can be applied to improve an arbitrary agreed-upon policy in a cooperative partially observable game. 
The first one, \emph{single-agent search}, effectively converts the problem into a single agent setting by making all but one of the agents play according to the agreed-upon policy.
In contrast, in \emph{multi-agent search} all agents carry out the same common-knowledge search procedure whenever doing so is computationally feasible, and fall back to playing according to the agreed-upon policy otherwise. 
We prove that these search procedures are theoretically guaranteed to at least maintain the original performance of the agreed-upon policy (up to a bounded approximation error).
In the benchmark challenge problem of Hanabi, our search technique greatly improves the performance of every agent we tested and when applied to a policy trained using RL achieves a new state-of-the-art score of 24.61 / 25 in the game, compared to a previous-best of 24.08 / 25.
\end{abstract}
\section{Introduction}
\label{sec:intro}

Real-world situations such as driving require humans to coordinate with others in a partially-observable environment with limited communication.
In such environments, humans have a mental model of how other agents will behave in different situations (theory of mind). This model allows them to change their beliefs about the world based on why they think an agent acted as they did, as well as predict how their own actions will affect others' future behavior. Together, these capabilities allow humans to search for a good action to take while accounting for the behavior of others.

Despite the importance of these cooperative settings to real-world applications, most recent progress on AI in large-scale games has been restricted to zero-sum settings where agents compete against each other, typically rendering communication useless. Search has been a key component in reaching professional-level performance in zero-sum games, including backgammon~\cite{tesauro1994td}, chess~\cite{campbell2002deep}, Go~\cite{silver2016mastering,silver2017mastering,silver2018general}, and poker~\cite{moravvcik2017deepstack,brown2017superhuman,brown2019superhuman}.

Inspired by the success of search techniques in these zero-sum settings,
in this paper we propose methods for agents to conduct search given an agreed-upon `convention' policy (which we call the \emph{blueprint policy}) in cooperative partially observable games. We refer to these techniques collectively as \emph{Search for Partially Observing Teams of Agents (SPARTA)}. In the first method, a single agent performs search assuming all other agents play according to the blueprint policy. This allows the search agent to treat the known policy of other agents as part of the environment and maintain beliefs about the hidden information based on others' actions.

In the second method, multiple agents can perform search simultaneously but must simulate the search procedure of other agents in order to understand why they took the actions they did. We propose a modification to the multi-agent search procedure - \emph{retrospective belief updates} - that allows agents to fall back to the blueprint policy when it is too expensive to compute their beliefs, which can drastically reduce the amount of computation while allowing for multi-agent search in most situations.

Going from just the blueprint policy, to single-agent search, to multi-agent search each empirically improves performance at the cost of increased computation. Additionally, we prove that SPARTA cannot result in a lower expected value than the blueprint policy except for an error term that decays in the number of Monte Carlo (MC) rollouts.

We test these techniques in Hanabi, which has been proposed as a new benchmark challenge problem for AI research~\cite{bard2019hanabi}. Hanabi is a popular fully cooperative, partially observable card game with limited communication. Most prior agents for Hanabi have been developed using handcrafted algorithms or deep reinforcement learning (RL).
However, the Hanabi challenge paper itself remarks that ``humans approach Hanabi differently than current learning algorithms'' because while RL algorithms perform exploration to find a good joint policy (convention), humans instead typically start with a convention and then individually search for the best action assuming that their partners will play the convention \cite{bard2019hanabi}. 

Applying SPARTA to an RL blueprint (that we train in self-play) establishes a new state-of-the-art score of 24.61~/~25 on 2-player Hanabi, compared to a previous-best of 24.08~/~25. Our search methods also achieve state-of-the-art scores in 3, 4, and 5-player Hanabi (Table~\ref{tab:search}). To our knowledge, this is the first application of theoretically sound search in a large partially observable cooperative game.

We provide code for single- and multi-agent search in Hanabi as well as a link to supplementary material at \url{https://github.com/facebookresearch/Hanabi_SPARTA}

\section{Related Work}
\label{sec:related_work}

Search has been necessary to achieve superhuman performance in almost every benchmark game. For fully observable games, examples include two-ply search in backgammon~\cite{tesauro1994td}, alpha-beta pruning in chess~\cite{campbell2002deep}, and Monte Carlo tree search in Go~\cite{silver2016mastering,silver2017mastering,silver2018general}. The most prominent partially observable benchmark game is poker, where search based on beliefs was key to achieving professional-level performance~\cite{moravvcik2017deepstack,brown2017superhuman,brown2019superhuman}.
Our search technique most closely resembles the one used in the superhuman multi-player poker bot \emph{Pluribus}, which conducts search given each agents' presumed beliefs and conducts MC rollouts beyond the depth limit of the search space assuming all agents play one of a small number of blueprint policies.

Cooperative multi-agent settings have been studied extensively under the DEC-POMDP formalism \cite{oliehoek2008optimal}. Finding optimal policies in DEC-POMDPs is known to be NEXP-hard \cite{bernstein2002complexity}, so much prior work studies environments with structure such as factorized local interactions \cite{oliehoek2008exploiting}, hierarchical policies \cite{amato2015planning}, or settings with explicit costs of communication \cite{goldman2003optimizing}.



There has been a good deal of prior work developing agents in Hanabi.
Notable examples of hand-crafted bots that incorporate human conventions include SmartBot~\cite{quuxplusone} and Fireflower~\cite{FireFlower}. Alternatively, so called `hat-coding' strategies~\cite{wtfwt}, which are particularly successful for N $>$ 2 players, instead use information theory by communicating instructions to all players via hints using modulo-coding, as is commonly employed to solve `hat-puzzles'.
More recent work has focused on tackling Hanabi as a learning problem~\cite{bard2019hanabi,foerster2019bayesian}. In this domain, the Bayesian Action Decoder learning method uses a public belief over private features and explores in the space of deterministic partial policies using deep RL~\cite{foerster2019bayesian}, which can be regarded as a scalable instantiation of the general ideas presented in~\cite{nayyar2013decentralized}. The Simplified Action Decoder algorithm established the most recent state of the art in Hanabi~\cite{anonymous2020simplified}.

Lastly, there is recent work on ad-hoc team play in Hanabi, in which agents get evaluated against a pool of different teammates~\cite{canaan2019diverse,walton2017evaluating}.

\section{Background}


We consider a Dec-POMDP
with $N$ agents. The Markov state of the environment is $s \in \mathcal{S}$ and we use $i$ to denote the agent index. At each time step, $t$, each agent obtains an observation $o^i = Z(s, i)$, where $Z$ is a deterministic observation function, and takes an action $a^i \in \mathcal{A}$, upon which the environment carries out a state transition based on the transition function, $s_{t+1} \sim \mathcal{P}(s_{t+1} | s_t, \mathbf{a})$, and agents receive a team-reward of $r_t=\mathcal{R}(s_{t},\mathbf{a})$, where $\mathbf{a}$ indicates the joint action of all agents. 
We use $\tau_t = \{s_0, \mathbf{a}_0, r_0 , ...s_t \}$ to denote the game history (or `trajectory') at time $t$.
Each agent's policy, $\pi^i$,  conditions on the agent's \emph{action-observation history} (AOH), $\tau^i_t = \{o^i_0, a^i_0, r_0 , ...o^i_t \}, $ and the goal of the agents is to maximise the total expected return, $J_{\pi} = \mathbb{E}_{\tau \sim P( \tau | \pi)} R_0(\tau)$, where $R_0(\tau)$ is the forward looking return of the trajectory, $R_t(\tau) = \sum_{t'\ge t} \gamma^{t' - t} r_{t'}$, and $\gamma$ is an (optional) discount factor. 
In this paper we consider both deterministic policies, $a^i = \pi(\tau^i_t)$, typically based on heuristics or search, and stochastic policies, $a^i \sim \pi^\theta(a^i |\tau^i_t)$, where $\theta$ are the parameters of a function approximator, \emph{e.g.}, a deep neural network.

In order to represent variable sequence lengths trajectories, $\tau^i_t$, Deep RL in partially observable settings typically uses recurrent networks (RNNs) ~\cite{hausknecht2015deep}. These RNNs can learn implicit representations of the sufficient statistics over the Markov state given $\tau^i_t$. In contrast, in our work we will use explicit \emph{beliefs} to represent the probability distribution over possible trajectories\footnote{Maintaining exact beliefs in large POMDPs is typically considered intractable. However we find that even in an environment with large state spaces such as card games, the number of states that have non-zero probability conditional on a player's observations is often much smaller; in the case of Hanabi, this set of possible states can be stored and updated explicitly.}. The private belief of agent $i$ that they are in trajectory $\tau_t$ at time step $t$ is $B^i(\tau_t) = P(\tau_t | \tau^i_t)$. We also define $B(\tau_t) = P(\tau_t | \tau_t^\mathcal{G})$ and $B(\tau^i_t) = P(\tau^i_t | \tau_t^\mathcal{G}) = \sum_{\tau_t \in \tau_t^i} B(\tau_t)$, which is the probability that agent~$i$ is in AOH $\tau^i_t$ conditional only on the \emph{common knowledge} (CK) of the entire group of agents $\tau^\mathcal{G}_t$.
Here CK are things that all agents know that all agents know ad infinitum. Please see~\cite{osborne1994course} for a formal definition of CK and \cite{foerster2018multi,foerster2019bayesian} for examples of how CK can arise and be used in multi-agent learning. Practically it can be computationally challenging to exactly compute common-knowledge beliefs due a large number of possible states and trajectories. However, in many settings, \emph{e.g.}, poker~\cite{brown2019superhuman}, the CK-belief can be factorized across public features and a set of private features associated with the different agents. In these settings the CK trajectory is simply the history of public features.


In our methods we will commonly need to carry out computations over each of the possible trajectories that have non-zero probability given an agent's AOH or all agents' common knowledge. We refer to these as the \emph{trajectory range} $\beta^i_t = \{\tau_t | B^i(\tau_t) > 0\}$ and $\beta_t = \{\tau_t | B(\tau_t) > 0\}$. We refer to the vector of probabilities in the trajectory range on timestep $t$ by ${\bf B}_t^i$ and ${\bf B_t}$. We also commonly need to carry out computations over each of an agent's AOHs that have non-zero probability given the common knowledge of the entire group of agents, which we refer to as the \emph{AOH range} $\chi^i_t = \{\tau^i_t | B(\tau^i_t) > 0\}$. We refer to the entire vector of probabilities in the AOH range of agent~$i$ on timestep $t$ by ${\bf C}^i_t$.

We further define the typical conditional expectations (`value functions'),

\begin{align}
V_{\pi}(\tau_t)   & = \mathbb{E}_{\tau'_T \sim P(\tau'_T |\pi, \tau_t)} R_t(\tau'_T)  \\
Q_\pi(\tau_t, a^i)& = \mathbb{E}_{\tau'_T \sim P(\tau'_T |\pi, \tau_t, a^i)} R_t(\tau'_T)
\end{align}
Given the range defined above we also introduce expectations conditioned on the AOHs, $\tau^i_t$:

\begin{align}
     V_{\pi}(\tau^i_t) &= \sum_{\tau_t \in \beta^i_t} B^i(\tau_t) V_{\pi}(\tau_t) \\
     Q_\pi(\tau^i_t, a^i) &= \sum_{\tau_t \in \beta^i_t} B^i(\tau_t) Q_{\pi}(\tau_t, a^i) 
\end{align}
Even though the optimal policies in the fully cooperative setting are deterministic, we consider stochastic policies for the purpose of reinforcement learning.


We further assume that a deterministic \emph{blueprint} policy $\pi_b$, defining what each player should do for all possible trajectories, is common knowledge amongst all players.
Player~$i$'s portion of the blueprint policy is denoted $\pi^i_b$ and the portion of all players other than $i$ as $\pi_b^{-i}$. In some settings, when actually playing, players may choose to not play according to the blueprint and instead choose a different action determined via online search. In that case $\pi_b$ differs from $\pi$, which denotes the policy that is actually played.

In our setting all of the past actions taken by all agents and the observation functions of all players are common knowledge to all agents. As such, if an agent is known to be playing according to a given policy, each action taken by this agent introduces a belief update across all other agents and the public belief. Suppose agent $i$ has a current belief $B_{t-1}^i$ and next observes $(a_t^j, o_t^i)$, where we have broken up the observation to separate out the observed partner action. Then,

\begin{align}
    \label{eq:belief}
     B^i(\tau_t) &= P(\tau_t | \tau_t^i) = P(\tau_t | \tau_{t - 1}^i, o^i_t, a_t^j) \\
     &= \frac{B^i(\tau_{t-1})  \pi^j(a_t^j | \tau_{t-1}) P(o^i_t | \tau_{t-1}, a_t^j)}{\sum\limits_{\tau'_{t-1}}{B^i(\tau'_{t-1})  \pi^j(a_t^j | \tau'_{t-1}) P(o^i_t | \tau'_{t-1}, a_t^j)}} \label{eq:belief3}
\end{align}

In other words, the belief update given $(a_t^j, o_t^i)$ consists of two updates: one based on the partner's known policy $\pi^j$, and the other based on the dynamics of the environment. The common knowledge belief $B$ is updated in the same way, using the common-knowledge observation $\tau^\mathcal{G}$ rather than $\tau^i$.





\section{Method}
\label{sec:method}
In this section we describe SPARTA, our online search algorithm for cooperative partially-observable games.


\subsection{Single-Agent Search}
\label{sec:single}
We first consider the case in which only one agent conducts online search. We denote the searching agent as agent~$i$. Every other agent simply plays according to the blueprint policy (and we assume agent~$i$ knows all other agents play according to the blueprint).

\begin{figure*}[ht!]
\centering
\includegraphics[width=117mm]{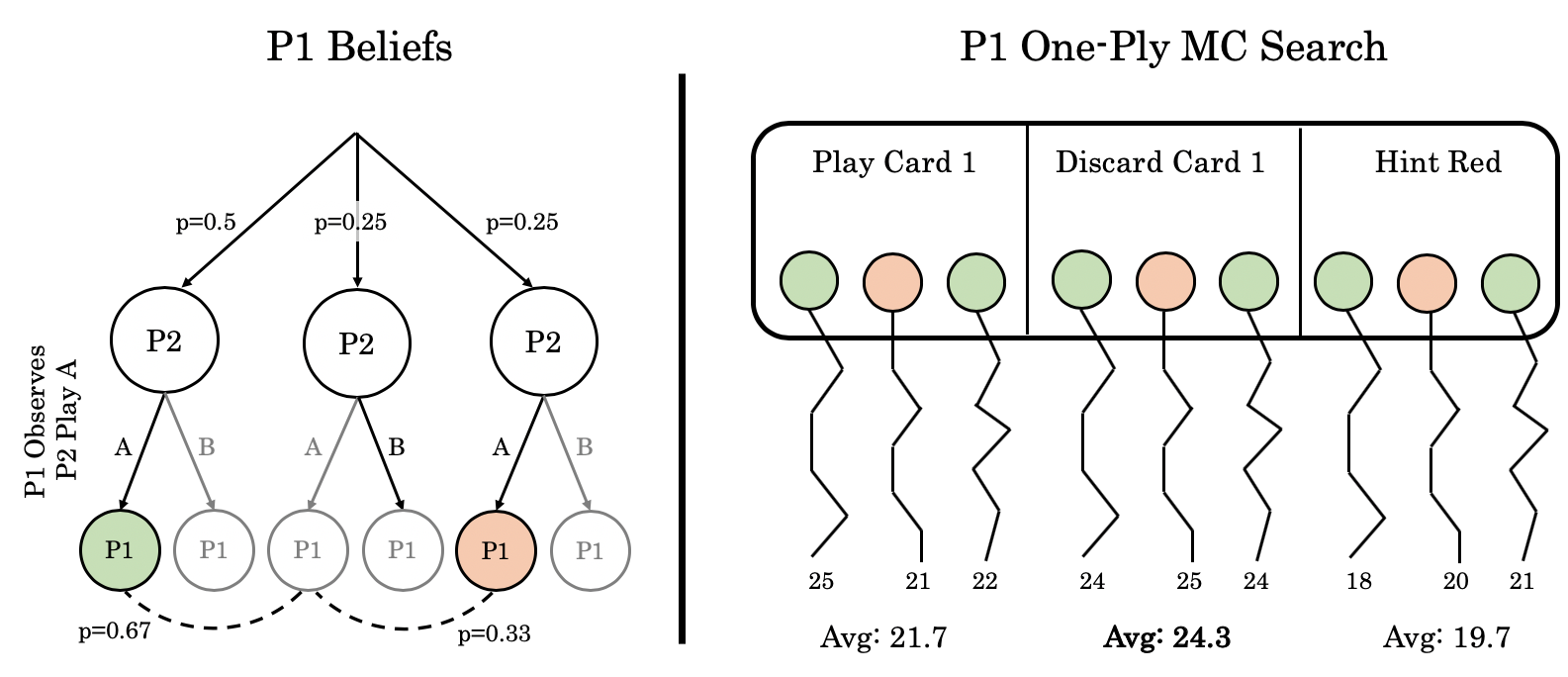}
\vspace{-3mm}
\caption{\small{An illustration of the SPARTA Monte Carlo search procedure for an agent P1. \textbf{Left:} The MDP starts in one of three states and P2 plays A. P1 observes that P2 played A, ruling out the middle state in which P2 would have played B. This leaves P1 with a probability distribution (beliefs) over two possible states. \textbf{Right:} For each legal action (`Play Card 1', ...) P1 performs rollouts from states drawn from the belief distribution, and picks the action with the highest average score: Discard Card 2.}}
\vspace{-0.05in}
\label{fig:hanabi_search}
\end{figure*}

Since agent~$i$ is the only agent determining her policy online while all other agents play a fixed common-knowledge policy, this is effectively a single-agent POMDP for agent~$i$. Specifically, agent~$i$ maintains a belief distribution ${\bf B}_t^i$ over trajectories she might be in based on her AOH $\tau_t^i$ and the known blueprint policy of the other agents. Each time she receives an observation or another agent acts, agent~$i$ updates her belief distribution according to (\ref{eq:belief3}) (see Figure \ref{fig:hanabi_search}, left). Each time $i$ must act, she estimates via Monte Carlo rollouts the expected value $Q_{\pi_b}(\tau_t^i, a^i)$ of each action assuming \emph{all} agents (including agent~$i$) play according to the joint blueprint policy $\pi_b$ for the remainder of the game following the action (Figure \ref{fig:hanabi_search}, right). A precise description of the single-agent search algorithm is provided in the appendix.


As described, agent~$i$ calculates the expected value of only the next action. This is referred to as 1-ply search. One could achieve even better performance by searching further ahead, or by having the agent choose between multiple blueprint policies for the remainder of the game. However, the computational cost of the search would also increase, especially in a game like Hanabi that has a large branching factor due to chance. In this paper all the experiments use 1-ply search.


Since this search procedure uses exact knowledge of all other agents' policies, it cannot be conducted correctly by multiple agents independently. That is, if agent~$j$ conducts search on a turn after agent~$i$ conducted search on a previous turn, then agent $j$'s beliefs are incorrect because they assume agent $i$ played $\pi^i_b$ while agent $i$ actually played the modified policy $\pi^i$. If more than one agent independently performs search assuming that others follow the blueprint, policy improvement cannot be guaranteed, and empirical performance is poor (Table \ref{tab:independent_search}, Appendix).

\subsection{Multi-Agent Search}
\label{sec:multi}

In order for an agent to conduct search effectively, her belief distribution must be accurate. In the case of single-agent search, this was achieved by all agents agreeing beforehand on a blueprint policy, and then also agreeing that only one agent would ever conduct search and deviate from the blueprint.
In this section we instead assume that all agents agree beforehand on both a blueprint policy and on what search procedure will be used.
When agent~$i$ acts and conducts search, the other agents exactly replicate the search procedure conducted by agent~$i$ (including the random seed) and compute agent~$i$'s resulting policy accordingly. In this way, the policy played so far is always common knowledge. 

Since the other agents do not know agent~$i$'s private observations, 
we have all agents (including agent~$i$) conduct search and compute agent~$i$'s policy for \emph{every possible} AOH that agent~$i$ might be in based on the common-knowledge observations.
Specifically, all agents conduct search for every AOH $\tau'^i_t \in \chi^i_t$. When conducting search for a particular $\tau'^i_t$ as part of this loop, the agents also compute what ${\bf B}_i^t$ would be assuming agent~$i$'s AOH is $\tau'^i_t$ and compute $Q_{\pi_b}(\tau'^i_t,a^i)$ for every action $a^i$ based on this ${\bf B}_i^t$. We refer to this loop of search over all $\tau'^i_t \in \chi_t^i$ as \emph{range-search}.

Agent~$i$ must also compute her policy via search for every $\tau'^i_t \in \chi^i_t$ (that is, conduct range-search) even though she knows $\tau^i_t$, because the search procedure of other agents on future timesteps may be based on agent~$i$'s policy for $\tau'^i_t \ne \tau^i_t$ where $\tau'^i_t \in \chi^i_t$ and it is necessary for all agents to be consistent on what that policy is to ensure that future search procedures are replicated identically by all agents.




In a game like two-player Hanabi, $|\chi^i_t|$ could be nearly 10 million, which means the range-search operation of multi-agent search could be 10 million times more expensive than single-agent search in some situations and therefore infeasible. Fortunately, the actual number of positive-probability AOHs will usually not be this large. We therefore have all agents agree beforehand on a budget for range-search, which we refer to as a \emph{max range} (abbreviated \textit{MR}). If $|\chi^i_t| > \textit{MR}$ on timestep $t$ where agent~$i$ is the acting agent, then agent~$i$ does not conduct search and instead simply plays according to the blueprint policy $\pi_b$. Since $\chi^i_t$ and the max range are common knowledge, it is also common knowledge when an agent does not search on a timestep and instead plays according to the blueprint.

Using a max range makes multi-agent search feasible on certain timesteps, but the fraction of timesteps in which search can be conducted may be less than single-agent search (which in a balanced two-player game is 50\%). The next section describes a way to use a max range while guaranteeing that there's always at least one agent who can perform search.

\subsection{Retrospective Belief Updates}

As discussed in the previous section, $\chi^i_t$ on some timesteps may be too large to conduct search on. Using a max range mitigates this problem, but may result in search only rarely being conducted. Fortunately, in many domains more common knowledge information is revealed as the game progresses, which reduces the number of positive-probability AOHs on previous timesteps.

For example, at the start of a two-player game of Hanabi in which agent~$i$ acts first (and then agent~$j$), $|\chi^i_0|$ might be nearly 10 million. If agent~$i$ were to use search to choose an action at this point, it would be too expensive for the agents to run range-search given the magnitude of $|\chi^i_0|$, so the other agents would not be able to conduct search on future timesteps.

However, suppose the action chosen from agent~$i$'s search results in agent~$i$ giving a hint to agent~$j$. Given this new common-knowledge information, it might now be known that only 100,000 of the 10 million seemingly possible AOHs at the first timestep were actually possible. It may now be feasible for the agents to run range-search on this reduced set of 100,000 possible AOHs. In this way, agent~$i$ is able to conduct search on a timestep $t$ where the max range is exceeded, and the agents can execute range-search at some later timestep $t'$ once further observations have reduced the size of $\chi^i_t$ below the max range.

We now introduce additional notation to generalize this idea. Agent~$i$'s belief at timestep $t'$ that the trajectory at some earlier timestep $t$ was (or is) $\tau_t$ is $B^i_{t'}(\tau_{t}) = P(\tau_t | \tau_{t'}^i)$. The public belief at timestep $t'$, which conditions only on the common knowledge of the entire group of agents at timestep $t'$, that the trajectory at timestep $t$ was (or is) $\tau_{t}$ is $B_{t'}(\tau_t) = P(\tau_t | \tau_{t'}^\mathcal{G})$ and $B_{t'}(\tau^i_t) = P(\tau^i_t | \tau_{t'}^\mathcal{G}) = \sum_{\tau_t \in \tau_t^i} B_{t'}(\tau_t)$. The trajectory range at timestep $t'$ of the trajectories at timestep $t$ is $\beta_{t,t'} = \{\tau_t | B_{t'}(\tau_t) > 0\}$ and $\beta^i_{t,t'} = \{\tau_t | B_{t'}^i(\tau_t) > 0\}$. The AOH range at timestep $t'$ of the agent~$i$ AOHs at timestep $t$ is $\chi^i_{t,t'} = \{\tau^i_t | B_{t'}(\tau^i_t) > 0\}$.

Again, the key idea behind retropective updates is that an agent can delay running range-search on a timestep until that range shrinks based on subsequent observations.
When it is an agent's turn to act, she conducts search if and only if she has run range-search for each previous timestep $t$ where one of the \emph{other} agents played search (because this means she knows her belief distribution). Otherwise, she plays according to the blueprint policy.

Specifically, all agents track the oldest timestep on which search was conducted but range-search was not conducted. This is denoted $t^*$. Assume the agent acting at $t^*$ was agent~$i$. If at any point $|\chi^i_{t^*,t}| \le \textit{MR}$ then all agents conduct range-search for timestep $t^*$ and $t^*$ is incremented up to the next timestep on which search was conducted but range-search was not conducted (but obviously not incremented past the current timestep $t$). If $|\chi^i_{t^*,t}| \le \textit{MR}$ for this new $t^*$ then all agents again conduct range-search and the process repeats. Since $\chi^i_{t^*,t}$ depends only on common knowledge, all agents conduct range-search at the same time and therefore $t^*$ is always consistent across all agents. When it is an agent's turn to act on timestep $t$, she conducts search if and only if she was the agent to act on timestep $t^*$ or if $t = t^*$.
An important upshot of this method is that it's always possible for at least one agent to use search when acting.

If $\textit{MR}$ is set to zero then multi-agent search with retrospective updates is identical to single-agent search, because the first agent to act in the game will conduct search and she will continue to be the only agent able to conduct search on future timesteps.
As $\textit{MR}$ is increased, agents are able to conduct search on an increasing fraction of timesteps.
In two-player Hanabi, setting $\textit{MR} =$ 10,000 makes it possible to conduct search on 86.0\% of timesteps even though in the worst case $|\chi^i_t| \approx$ 10,000,000.


\subsection{Soundness and Convergence Bound for Search}

We now prove a theorem that applies to all three SPARTA variants described in this section. Loosely, it states that applying search on top of a blueprint policy cannot reduce the expected reward relative to the blueprint, except for an error term that decays as $O(1/\sqrt{N})$, where $N$ is the number of Monte Carlo rollouts.

\begin{theorem}
Consider a Dec-POMDP with N agents, actions $\mathcal{A}$, reward bounded by $r_{\textit{min}} \leq R(\tau) < r_{\textit{max}}$ where $r_{\textit{max}} - r_{\textit{min}} \leq \Delta$, game length bounded by $T$, and a set of blueprint policies $\pi_b \equiv \{\pi_b^0,\ldots,\pi_b^N\}$. If a Monte Carlo search policy $\pi_s$ is applied using $N$ rollouts per step, then

\begin{equation}
    V_{\pi_s} - V_{\pi_b} \geq -2T\Delta |\mathcal{A}|N^{-1/2}
\end{equation}

\end{theorem}

The proof is provided in the Appendix.

\section{Experimental Setup}


We evaluate our methods in the partially observable, fully cooperative game Hanabi, which at a high level resembles a cooperative extension of solitaire. 
Hanabi has recently been proposed as a new frontier for AI research~\cite{bard2019hanabi} with a unique focus on theory of mind and communication.

The main goal of the team of agents is to complete 5 stacks of cards, one for each color, in a legal sequence, starting with a $\mathbf{1}$ and finishing with a $\mathbf{5}$.
The defining twist in Hanabi is that while players can observe the cards held by their teammates, they cannot observe their own cards. As such, players need to exchange information with their teammates in order to decide which cards to play. Hanabi offers two different means for doing so. First, players can take costly hint actions that reveal part of the state to their teammates.
Second, since all actions are observed by all players, each action (such as playing a card or discarding a card) can itself be used to convey information, in particular if players agree on a set of conventions before the game. For further details on the state and action space in Hanabi please see~\cite{bard2019hanabi}.

In this work we are focused on the \emph{self-play} part of the challenge, in which the goal is to find a set of policies that achieve a high score when playing together as a team.

For the blueprint strategy used in the experiments, we experimented with open-sourced handcrafted bots WTFWThat~\cite{wtfwt} and SmartBot~\cite{quuxplusone}. We also created two of our own blueprint strategies. One was generated from scratch using deep reinforcement learning, which we call RLBot. The other, which we call CloneBot, was generated by conducting imitation learning using deep neural networks on the policy produced by single-agent search on top of SmartBot. The details for the generation of both bots are given in the appendix.

All experiments except the imitation learning of CloneBot and the reinforcement learning of RLBot were conducted on CPU using machines with Intel\textsuperscript{\textregistered} Xeon\textsuperscript{\textregistered} E5-2698 CPUs containing 40 cores each. A game of Hanabi requires about 2 core-hours for single-agent search and 90 core-hours for retrospective multi-agent search using the SmartBot blueprint policy with a max range of 10,000. We parallelize the search procedure over multiple cores on a single machine.

\subsection{Implementing SPARTA for Hanabi}

The public and private observations in Hanabi are factorizable into public and private features.\footnote{The class of games of this form has recently been formalized as `Factorized Observation Games' in \cite{kovavrik2019rethinking}.} Specifically, each player's hidden information consists of the cards held by other agents, so the beliefs for each player can be represented as a distribution over possible hands that player may be holding. The initial private beliefs can be constructed based on the card counts. In addition to updates based on the actions of other agents, updates based on observations amount to (a) adjusting the probabilities of each hand based on the modified card count as cards are revealed, and (b) setting the probability of hands inconsistent with hints to 0.

The public beliefs for 2-player Hanabi can be factored into independent probability distributions over each player's hand (conditional on the common knowledge observations). These public beliefs are identical to the private beliefs except that the card counts are not adjusted for the cards in the partner's hand. Given one player's hand, the private beliefs over the other player's hand (which is of course no longer independent of the other player's hand) can be computed from the public beliefs by adjusting the card counts for the privately-observed cards.\footnote{This conversion from public to private beliefs is what we call ConditionOnAOH() in the algorithm listing in the Appendix.}



\subsection{Estimating Action Expected Values via UCB}

In order to reduce the number of MC rollouts that must be performed during search, we use a UCB-like procedure that skips MC rollouts for actions that are presumed not to be the highest-value action with high confidence.
After a minimum of 100 rollouts per action is performed, the reward sample mean and its standard deviation is computed for each action. If the expected value for an action is not within 2 standard deviations of the expected value of the best action, its future MC rollouts are skipped.

Furthermore, we use a configurable threshold for deviating from the blueprint action. If the expected value of the action chosen by search does not exceed the value of the blueprint action by more than this threshold, the agent plays the blueprint action. We use a threshold of $0.05$ in our experiments.

The combination of UCB and the blueprint deviation threshold reduces the number of rollouts required per timestep by 10$\times$, as shown in Figure~\ref{fig:search_sweep} in the appendix.

\subsection{Bootstrapping Search-Based Policies via Imitation Learning}

Search can be thought of as a policy improvement operator, i.e. an algorithm that takes in a joint policy and outputs samples from an improved joint policy. These samples can be used as training data to learn a new policy via imitation learning. This improved policy approximates the effect of search on the blueprint policy while being cheaper to execute, and search can be run on this learned policy, in effect bootstrapping the search procedure. In principle, repeated application of search and learning could allow the effect of single-agent search to be applied on multiple agents, and could allow the benefits of search to extend beyond the depth limit of the search procedure. However, there is no guarantee that this process would eventually converge to an optimal policy, even in the case of perfect function approximation.

We refer to the policy learned in this manner as CloneBot. We provide details of the training procedure in the Appendix and evaluate its performance in Table \ref{tab:search}.
\label{sec:supervised}
\section{Results}
\label{sec:results}

Table~\ref{tab:search} shows that adding SPARTA leads to a large improvement in performance for all blueprint policies tested in two-player Hanabi (the most challenging variant of Hanabi for computers) and achieves a new state-of-the-art score of 24.61 / 25 compared to the previous-best of 24.08 / 25.\footnote{It is impossible to achieve a perfect score for some shuffles of the deck. However, the highest possible average score is unknown.} Much of this improvement comes from adding single-agent search, though adding multi-agent search leads to an additional substantial improvement. 

Unfortunately there are no reliable statistics on top human performance in Hanabi. Discussions with highly experienced human players has suggested that top players might achieve perfect scores in 2-player Hanabi somewhere in the range of 60\% to 70\% of the time when optimizing for perfect scores. Our strongest agent optimizes for expected value rather than perfect scores and still achieves perfect scores 75.5\% of the time in 2-player Hanabi.

\begin{table*}[h!]
\begin{center}
\begin{tabular}{ l | c c c }
\toprule
{\bf Blueprint Strategy} & {\bf No Search} & {\bf Single-Agent Search} & {\bf Multi-Agent Search} \\
\midrule
RL (ACHA)  & 22.73 $\pm$ 0.12 & - & - \\
\textit{\cite{bard2019hanabi}} & 15.1\% & - & - \\
\midrule

BAD  & 23.92 $\pm$ 0.01 & - & - \\
\textit{\cite{foerster2019bayesian}} & 58.6\% & - & -\\
\midrule
\midrule

SmartBot  & 22.99 $\pm$ 0.001 & {\bf 24.21 $\pm$ 0.002} & {\bf 24.46 $\pm$ 0.01}  \\
\textit{\cite{quuxplusone}}& 29.6\% & {\bf 59.1\%} & {\bf 64.8\%} \\
\midrule
CloneBot & 23.32 $\pm$ 0.001 & {\bf 24.37 $\pm$ 0.02} & - \\
{\it (ours)} & 40.6\% & {\bf 64.6\%} & - \\
\midrule
DQN & 23.45 $\pm$ 0.01 & {\bf 24.30 $\pm$ 0.02} & {\bf 24.49 $\pm$ 0.02} \\
{\it (ours)} & 46.8\% & {\bf 63.5\%} & {\bf 67.7\%} \\
\midrule
SAD & 24.08 $\pm$ 0.01 & {\bf 24.53 $\pm$ 0.01} & {\bf 24.61 $\pm$ 0.01} \\
{\it \cite{anonymous2020simplified}} & 56.1\% & {\bf 71.1\%} & {\bf 75.5\%} \\
\bottomrule
\end{tabular}
\end{center}
\caption{\small Results for various policies in 2-player Hanabi coupled with single-agent and retrospective multi-agent search (SPARTA). The top row is the average score (25 is perfect), the bottom row is the perfect-score percentage. Single-agent and multi-agent search monotonically improve performance. For multi-agent search, a max range of 2,000 is used. Results in bold beat the prior state of the art of 24.08.
}
\label{tab:search}
\end{table*}

Table~\ref{tab:sad_kplayer} shows the benefits of single-agent search also extend to the 3, 4, and 5-player variants of Hanabi as well. For these variants, the \textit{SAD} agent~\cite{anonymous2020simplified} was state-of-the-art among learned policies, while an information-theoretic hat-coding policy (WTFWThat) achieves close to perfect scores for 4 and 5 players~\cite{wtfwt}. Applying single-agent search to either SAD or WTFWThat improves the state-of-the-art scores for 3, 4, and 5-player variants (Table~\ref{tab:sad_kplayer}, Table~\ref{tab:wtfwthat} in Appendix). Applying multi-agent search in Hanabi becomes much more expensive as the number of players grows because the number of cards each player observes becomes larger.

\begin{table}[h]
\begin{center}
\begin{tabular}{ c | c c }
\toprule
{\bf \# Players} & {\bf No Search} & {\bf Single-Agent Search} \\
\midrule
\multirow{2}{*}{2} & 24.08 $\pm$ 0.01 & {\bf 24.53 $\pm$ 0.01} \\
  & 56.1\% & \textbf{71.1\%} \\
\midrule
\multirow{2}{*}{3} & 23.99 $\pm$ 0.01 & {\bf 24.61 $\pm$ 0.01}
\\
  & 50.4\% & {\bf 74.5\%} \\
\midrule
\multirow{2}{*}{4} & 23.81 $\pm$ 0.01 & {\bf 24.49 $\pm$ 0.01}
\\
  & 41.5\% & {\bf 64.2\%} \\
\midrule
\multirow{2}{*}{5} & 23.01 $\pm$ 0.01 & {\bf 23.91 $\pm$ 0.01}
\\
  & 13.9\% & {\bf 32.9\%} \\
\bottomrule
\end{tabular}
\end{center}

\caption{\small Hanabi performance for different numbers of players, applying single-agent search (SPARTA) to the `SAD' blueprint agent which achieved state-of-the-art performance in 3, 4, and 5-player Hanabi among learned policies. Results shown in bold are state-of-the-art (for learned policies). Higher scores in 3+ player Hanabi can be achieved using hard-coded `hat-counting` techniques; Table \ref{tab:wtfwthat} in the Appendix shows that search improves those strategies too.}
\label{tab:sad_kplayer}
\end{table}

Table~\ref{tab:maxrange} examines the performance and cost in number of rollouts for single-agent search and multi-agent search with different values of the max range (MR) parameter, using SmartBot as the blueprint agent. When MR is set to zero, we are performing single-agent search, so search is conducted on 50\% of timesteps, which requires about $10^5$ rollouts per game\footnote{SPARTA also performs counterfactual belief updates using the blueprint, using about $2\times 10^6$ policy evaluations per game, which corresponds to about $4\times 10^4$ games worth of policy evaluations. This cost is dominated by search rollouts for all SPARTA variants.}. As the max range is increased, search is conducted more often. At a max range of 10,000, search is conducted on 86\% of timesteps even though the maximum possible range for a timestep in two-player Hanabi is nearly 10 million. However, this still requires about 1,000$\times$ as many rollouts as single-agent search.

\begin{table}[h]
\begin{center}
\begin{tabular}{ c | c c c }
\toprule
{\bf Max Range} & {\bf \% search} & {\bf \# Rollouts} & {\bf Average Score} \\
\midrule
0      & 50.0\% & $1.5 \times 10^5$ & 24.21 $\pm$ 0.002 \\
80     & 56.2\% & $3.0 \times 10^6$ & 24.30 $\pm$ 0.02 \\
400    & 65.1\% & $1.3 \times 10^7$ & 24.40 $\pm$ 0.02 \\
2000   & 77.2\% & $5.2 \times 10^7$ & 24.46 $\pm$ 0.01 \\
10000  & 86.0\% & $1.8 \times 10^8$ & 24.48 $\pm$ 0.01 \\
\bottomrule
\end{tabular}
\end{center}

\caption{\small Multi-agent retrospective search with a SmartBot blueprint for different values of max range. As max range increases, search is performed more often leading to a higher score, but the rollout cost increases dramatically.}

\label{tab:maxrange}
\end{table}


Figure \ref{fig:ev} plots the average MC search prediction of the expected payoff of the best move ($Q_{\pi_b}(\tau^i, a^*)$) at different points in the game, for two blueprint policies. As predicted by the theory, the expected score starts at the blueprint expected score and increases monotonically as search is applied at each move of the game.

\begin{figure}[t]
\centering
\includegraphics[width=\columnwidth]{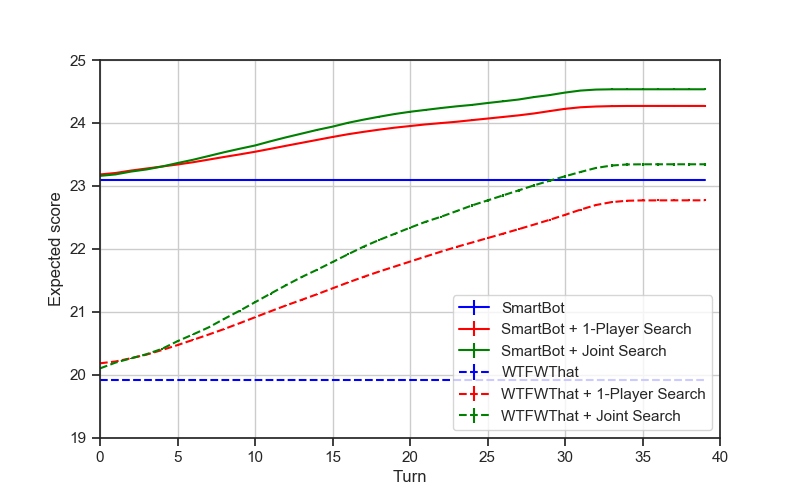}
\label{fig:ev}
\caption{\small{Average score predicted by MC search rollouts at different points in the game, for two different blueprints. Expected score by turn is averaged over replicates for each condition (error bars are included but are too small to be visible). If a game ends before 40 turns, the final score is propagated for all subsequent turns. }}
\label{fig:ev}
\end{figure}
\section{Conclusions}

In this paper we described approaches to search in partially observable cooperative games that improves upon an arbitrary blueprint policy. Our algorithms ensure that any agent conducting search always has an accurate belief distribution over the possible trajectories they may be in, and provide an alternative in case the search procedure is intractable at certain timesteps. We showed that in the benchmark domain of Hanabi, both search techniques lead to large improvements in performance to all the policies we tested on, and achieves a new state of the art score of 24.61 / 25 compared to a previous-best of 24.08 / 25. We also proved that applying our search procedure cannot hurt the expected reward relative to the blueprint policy, except for an error term that shrinks with more Monte Carlo rollouts. This result fits a theme, also shown in other games such as Chess, Go, and Poker, that search leads to dramatically improved performance compared to learning or heuristics alone.

The performance improvements from search come at a computational cost. Our search procedure involves tracking the belief probability of each AOH that a player may be in given the public information, which for Hanabi is no more than 10 million probabilities. Search also requires computing a large number of MC rollouts of the policy, especially for multi-agent search where the number of rollouts scales with the size of the belief space.  Conducting search in partially observable games, whether cooperative or competitive, with many more AOHs per instance of public information remains an interesting challenge for future work and may be applicable to settings like Bridge and environments with visual inputs like driving. 

This work currently assumes perfect knowledge of other agents' policies. This is a reasonable assumption in settings involving centralized planning but decentralized execution, such as self-driving cars that are created by a single company or robots working together as a team in a factory. In general however, an agent's model of others may not be perfect. Investigating how search performs in the presence of an imperfect model of the partner, and how to make search more robust to errors in that model, are important directions for future work as well.

\section{Acknowledgments}

We would like to thank Pratik Ringshia for developing user interfaces used to interact with Hanabi agents.


\bibliographystyle{aaai}
\bibliography{references}

\appendix
\clearpage

\clearpage
\onecolumn
\section{Search Algorithm Listing}
\label{app:algorithm}
\begin{algorithm}[!ht]
\caption{Single Agent Search}
\begin{algorithmic}
\Function{SingleAgentSearch}{$i$, $\pi_0 \ldots \pi_N$}
\State Initialize $B_0^i$, agent~$i$'s private beliefs.
\For{$t = 1,\ldots,T$}
    \State $o_t \gets \Call{GetEnvironmentObservation}$
    \State $B_t^i \gets \Call{StepBeliefs}{B_{t-1}^i, o_t}$
    \State $j \gets \Call{whoseTurn}{t}$
    \If{$j = i$} 
        \State $a_t \gets \Call{MCSearch}{B_t^i, i, \pi_0 \ldots \pi_N}$
        \State Play $a_t$
    \Else
        \State $a_t \gets \Call{GetTeammateAction}$ \Comment{Observe the teammate's action}
        \For{$(\tau, p_\tau) \in \beta_t^i$}  \Comment{$\beta_t^i$ are the non-zero elements of $B^i_t$} 
            \State $a_t' \gets \pi_j(\tau^j)$ \Comment{Assuming deterministic $\pi_j$}
            \State If $a_t' \neq a_t$, set $p_\tau$ to $0$.
        \EndFor
    \EndIf
\EndFor
\EndFunction
\\
\Function{MCSearch}{$B^i$, $i$, $\pi_0 \ldots \pi_N$}
    \State $\textrm{stats}[a] \gets 0$ for $a\in A$
    \While{not converged} \Comment{We don't describe UCB here}
        \State $\tau \sim B^i$ \Comment{Sample $\tau$ from the beliefs according to $p_{\tau}$}
        \For{$a \in A$}
            \State score $\gets$ \Call{RolloutGame}{$\tau$, $a$, $\pi_0 \ldots \pi_N$}
            \State $\textrm{stats}[a] \gets \textrm{stats}[a] + \textrm{score}$
        \EndFor
    \EndWhile
    \Return $\argmax\limits_{a\in A}$ stats[a]
\EndFunction
\\

\end{algorithmic}
\label{alg:deepcfrx}
\end{algorithm}

\label{app:algorithmx}
\begin{algorithm}[!ht]
\caption{Multi-Agent Search}
\begin{algorithmic}

\Function{RetrospectiveMultiAgentSearch}{$i, \pi_0, \ldots, \pi_N, maxRange$}
\State Initialize $B_0$, the common knowledge beliefs at $t=0$.
\State $searcher \gets 0$ \Comment{The player that is {\it currently} allowed to perform search}
\State $actionQ \gets \varnothing$ \Comment{The set of actions for which the beliefs have not been updated}
\For{$t = 1,\ldots,T$}
    \State $o_t \gets \Call{GetEnvironmentObservation}$
    \State $j \gets \Call{WhoseTurn}{t}$
    \State $B_t \gets \Call{StepBeliefs}{B_{t-1}, o_t^\mathcal{G}}$  \Comment{$o_t^\mathcal{G}$ is the common-knowledge observation}
    \State \Call{RetrospectiveBeliefUpdate}{$\pi_0, \ldots, \pi_N,o_t^\mathcal{G},actionQ,  maxRange$}
    \If{$actionQ = \varnothing$}
        \State $searcher = j$
    \EndIf
    \If{$j = i$}  
        \State $a_t \gets $\Call{SelectMove}{$B_t, i,  \pi_0, \ldots, \pi_N, searcher$}
        \State Play $a_t$
    \Else
        \State $a_t \gets \Call{GetTeammateAction}$ \Comment{Observe the teammate's action}
    \EndIf
    \State $\chi_t^j \gets $\Call{GetAohs}{$B_t, j$}
    \If {$j = searcher$} 
        \State $actionQ.push\_back((t, a_t, j, B_t, \chi_t^j))$ 
    \Else 
        \For{$(\tau^j, p_{\tau^j}) \in \chi_t^j$} \Comment{Update beliefs immediately based on blueprint $\pi_j$}
            \State $a_t' \gets \pi_j(\tau^j)$ \Comment{Assuming deterministic $\pi_j$}
            \State If $a_t' \neq a_t$, set $p_{\tau^j}$ to $0$.
        \EndFor
    \EndIf
\EndFor
\EndFunction
\\ 

\Function{SelectMove}{$B_t, i, \pi_0, \ldots, \pi_N, searcher$}
    \If{$i = searcher$}
        \State $B_t^i \gets \Call{ConditionOnAOH}{B_t, \tau^i}$
        \State $a \gets \Call{MCSearch}{B_t^i, i, \pi_0,\ldots,\pi_N}$
    \Else
        \State $a \gets \pi_i(\tau^i)$
    \EndIf
    \Return $a$
\EndFunction
\\

\Function{RetrospectiveBeliefUpdate}{$\pi_0, \ldots, \pi_N,o_t^\mathcal{G},actionQ, maxRange$}
    \For{$(a_{t'}, j, B_{t'}, \chi_{t',t}^j) \in actionQ$}
        \State Update $\chi_{t',t}^j$ based on $o_t^\mathcal{G}$
    \EndFor
    \While{$actionQ \neq \varnothing$}
        \State $(a_{t'}, j, B_{t'}, \chi_{t',t}^j) \gets actionQ.front()$
        \If{$|\chi_{t',t}^j| > maxRange$}
            \State break
        \EndIf
        \For{$(\tau^j, p_{\tau^j}) \in \chi_{t',t}^j$}
            \State $B_{t'}^j \gets \Call{ConditionOnAOH}{B_{t'}, \tau^j}$
            \State $a_{t'}' \gets \Call{MCSearch}{B_{t'}^j, j, \pi_0 \ldots \pi_N}$
            \If{$a_{t'}' \neq a_{t'}$}
                \For{$(a_{t'}, j, B_{t'}, \chi_{t',t}^j) \in actionQ$} \Comment{Prune beliefs at subsequent times}
                    \State Remove any $\tau\in \chi_{t',t}^j$ for which $\tau^j$ is a prefix of $\tau$   
                \EndFor
            \EndIf
        \EndFor
        \State $actionQ.pop\_front()$
    \EndWhile
\EndFunction

\end{algorithmic}
\label{alg:deepcfrx}
\end{algorithm}

\twocolumn
\clearpage
\section{Experimental Details for Reinforcement Learning}
\label{app:rl_training}

We train the reinforcement learning baseline with our own implementation of Ape-X DQN~\cite{apex} on the Hanabi Learning Environment (HLE)~\cite{bard2019hanabi}. Essentially, Ape-X DQN is a distributed Q-learning framework with a large number of asynchronous actors feeding transitions into a shared prioritized replay buffer~\cite{prioritized-replay} in parallel, and a centralized learner that samples from the replay buffer to update the agent. It also incorporates several techniques for improved performance such as $n$-step return targets~\cite{Sutton:1988:td}, double Q-learning~\cite{double-dqn} and dueling network architecture~\cite{dueling-dqn}. There are two notable differences between our implementation and the one proposed in the original paper. First, instead of having 360 actors each running on a single environment on a CPU, we use 80 actors while each of them running on a vector of 20 environments. We loop over 20 environments and batch their observations together. Then the actor acts on the batch data using GPU. With this modification, our implementation can run under moderate computation resources with 20 CPU cores (40 Hyper-Threads) and 2 GPUs where one GPUs is used for training and the other is shared by all 80 asynchronous actors. Second, we synchronize all actors with learner every 10 mini-batches while in the original paper each actor independently synchronize with the learner every 400 environment steps.

We modify the input features of HLE by replacing card knowledge section with the V0-Belief proposed in~\cite{foerster2019bayesian}. To avoid the agent being overly cautious, we do not zero out reward for games in which all life tokens are exhausted during training, even though we report numbers according to the counting scheme at test time. The agent uses a 3 layer fully connected network with 512 neurons each layer, followed by two-stream output layers for value and advantage respectively. Similar to the original Ape-X paper, each actor executes an $\epsilon_i$-greedy policy where $\epsilon_i = \epsilon^{1 + \frac{1}{N-1}\alpha}$ for $i \in \{0, ..., N-1\}$ but with a smaller $\epsilon=0.1$ and $\alpha=7$. The discount factor $\gamma$ is set to $0.999$. The agent is trained using Adam optimizer~\cite{kingma2014adam} with learning rate $=6.25 \times 10^{-5}$ and $\epsilon=1.5 \times 10^{-5}$. Each mini-batch contains $512$ transitions sampled from the prioritized replay buffer with priority exponent of $0.6$ and importance sampling exponent equal to $0.4$.

\section{Experimental Details for Imitation Learning Applied to Search}

\label{app:clonebot}
We train a supervised policy for one agent using sampled AOH-action pairs from 1.3 million games where single-agent search was applied to the SmartBot blueprint policy. The model is a 2-layer, 128-unit LSTM, which takes as input a representation of the AOH as described in the previous section, as well as the action the blueprint agent plays at this AOH. This latter input is available at inference time and is crucial to achieve good performance, presumably because the neural model can ``fall back'' to the blueprint policy, which may be hard to approximate in some situations (as it's not a neural policy). 

The model outputs a softmax policy $\pi_{clone}$ and is optimized using the objective function is the expected reward $\sum_a {\pi_{clone}(\tau^i, a) Q(\tau^i, a)}$ . At test time we execute the action with the highest probability under the policy. We found that this objective led to a stronger agent than either predicting $Q(\tau^i, a)$ directly with an MSE loss - which wastes model capacity predicting $Q$ for unplayed actions - or predicting $a$ directly with a cross-entropy loss - which forces the model to select between actions with nearly identical value.

We train the model for 100,000 SGD steps, a batch size of 256, using the RMSprop optimizer with a learning rate of $5\times 10^{-4}$.

\section{Additional Results}

\begin{figure}[t]
\centering
\includegraphics[width=\columnwidth]{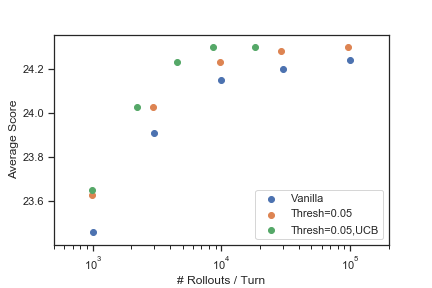}
\vspace{-5mm}
\caption{\small{Effect of blueprint deviation threshold and UCB-like search pruning on average score as a function of total number of rollouts per turn when using SmartBot as the blueprint. }}
\label{fig:search_sweep}
\end{figure}

Figure \ref{fig:search_sweep} shows the effect of the blueprint deviation threshold and UCB-like search pruning on average score. Interestingly, a threshold of 0.05 improves the total score even for a large number of rollouts; perhaps this is because the move chosen by SmartBot is more useful for future search optimizations later in the game. UCB reduces the number of rollouts per turn without affecting the average score.

Table \ref{tab:independent_search} lists average scores in 2-player Hanabi using \emph{independent multi-agent search}. In this variant, both players independently perform search (incorrectly) assuming that the partner is playing the blueprint. Without modification, this approach leads to undefined situations where the partner's beliefs contain no states. To solve this, we add some uncertainty to the model of the partner's strategy. 

In the table, we sweep two parameters: the uncertainty about the partner's strategy in the belief update, and the threshold minimum difference in expected reward at which the agent will deviate from the blueprint. The rationale for the threshold is that any time a player deviates from the blueprint, they are corrupting their partner's beliefs, so this should only be done when it is substantially beneficial. Under some settings, independent joint search performs better than the blueprint (22.99) but never outperforms single-agent search.

\begin{table}[h]
\begin{center}
\begin{tabular}{ c | c c c c }
\toprule
  {\bf Belief} & \multicolumn{4}{c}{\bf $\Delta R$ ~Threshold} \\
 {\bf Uncertainty} & 0 & 0.1 & 0.2 & 0.5 \\
\midrule
0.05 & 14.41 & 23.62 & 24.18 & 24.02 \\
0.1 & 14.22 & 23.48 & 24.12 & 24.01 \\
0.2 & 13.69 & 22.41 & 23.57 & 23.83 \\
\bottomrule
\end{tabular}
\end{center}
\caption{\small Average scores in 2-player Hanabi for \textit{independent multi-agent search} using the SmartBot blueprint strategy. }
\label{tab:independent_search}
\end{table}

Table \ref{tab:wtfwthat} shows new state-of-the-art results in 3, 4, and 5 player Hanabi by applying search on top of the information-theoretic `WTFWThat` hat-coding policy. This hat-coding policy achieves near-perfect scores but is not learned and considered somewhat against the spirit of the game, since it does not use grounded information at all. Nevertheless, we show here that even these policies can be improved with single-agent search.

\begin{table}[h]
\begin{center}
\begin{tabular}{ c | c c }
\toprule
{\bf \# Players} & {\bf No Search} & {\bf Single-Agent Search} \\
\midrule
\multirow{2}{*}{2} & 19.45 $\pm$ 0.001 & 22.78 $\pm$ 0.02 \\
  & 0.28\% & 10.0\% \\
\midrule
\multirow{2}{*}{3} & 24.20 $\pm$ 0.01 & {\bf 24.83 $\pm$ 0.006}
\\
  & 49.1\% & {\bf 85.9\%} \\
\midrule
\multirow{2}{*}{4} & 24.83 $\pm$ 0.01 & {\bf 24.96 $\pm$ 0.003}
\\
  & 87.2\% & {\bf 96.4\%} \\
\midrule
\multirow{2}{*}{5} & 24.89 $\pm$ 0.00 & {\bf 24.94 $\pm$ 0.004}
\\
  & 91.5\% & {\bf 95.5\%} \\
\bottomrule
\end{tabular}
\end{center}

\caption{\small Hanabi performance for different numbers of players, using the hard-coded `WTFWThat' hat-counting blueprint agent which achieved state-of-art performance in 3, 4, and 5-player Hanabi. Results shown in bold are state-of-the-art for that number of players.}
\label{tab:wtfwthat}
\end{table}
\onecolumn
\section{Proof of Main Theorem}

\vspace{24pt}

\label{app:theorem}
\newcommand{\pb}{\ensuremath{{\pi_b}}}

\newcommand{\E}[1]{\ensuremath{\mathbb{E}\left[#1\right]}}
\newcommand{\A}{\ensuremath{\mathcal{A}}}
\newcommand{\bR}{\ensuremath{Q_\pb(\tau^i,a)}}
\newcommand{\hR}{\ensuremath{\hat{Q}_\pb(\tau^i,a)}}

\begin{theorem}
Consider a Dec-POMDP with N agents, actions $\A$, reward bounded by $r_{\textit{min}} \leq R(\tau) < r_{\textit{max}}$ where $r_{\textit{max}} - r_{\textit{min}} \leq \Delta$, game length bounded by $T$, and a set of blueprint policies $\pi_b \equiv \{\pi_b^0 \ldots \pi_b^N\}$. If a MC search policy $\pi_s$ is applied using $N$ rollouts per step, then

\begin{equation}
    V_{\pi_s} - V_{\pi_b} \geq -2T\Delta |\A|N^{-1/2}
\end{equation}

\end{theorem}

\begin{proof}
Consider agent $i$ at some (true) trajectory $\tau$. This agent has an exact set of beliefs $B^i(\tau^i)$, i.e. the exact probability distribution over trajectories it could be in given its action-observation history $\tau^i$.

\vspace{8 pt}

Suppose that an agent at some point $\tau$ acts according to the search procedure described in Section \ref{sec:method} (it is not relevant whether single-agent or joint search is performed). For each action $a \in \A$, the agent collects $N/|\A|$ i.i.d. samples of the reward $R$, sampling trajectories from $\tau \sim B^i(\tau^i)$ and assuming that agent $i$ plays some action $a$ at $\tau^i$ and all agents play according to $\pb$ thereafter. We denote a single MC rollout (which is an MC estimate of \bR) as $\hat{Q}_\pb^k(\tau^i,a)$, and the mean of the rollouts for $a$ as \hR. The agent then plays $\hat{a}^* \equiv \argmax\limits_{a\in \A}{\hR}$. We denote the {\it true} optimal action as $a^* \equiv \argmax\limits_{a\in \A}{\bR}$. 

\begin{align}
\small
Q_\pb(\tau^i,\hat{a}^*) - V_\pb(\tau^i) &= Q_\pb(\tau^i,\hat{a}^*) - Q_\pb(\tau^i,\pb(\tau^i)) \\
&= Q_\pb(\tau^i,\hat{a}^*) - \hat{Q}_\pb(\tau^i,\hat{a}^*) + \hat{Q}_\pb(\tau^i,\hat{a}^*) \label{eq:q2}\\
& \hspace{1cm} -\hat{Q}_\pb(\tau^i,\pb(\tau^i)) + \hat{Q}_\pb(\tau^i,\pb(\tau^i)) - Q_\pb(\tau^i,\pb(\tau^i)) \nonumber \\
& \ge Q_\pb(\tau^i,\hat{a}^*) - \hat{Q}_\pb(\tau^i,\hat{a}^*) +  \hat{Q}_\pb(\tau^i,\pb(\tau^i)) - Q_\pb(\tau^i,\pb(\tau^i)) \label{eq:q3}\\
& \ge -2~\max\limits_{a\in \A}~\left| \hat{Q}_\pb(\tau^i,a) - Q_\pb(\tau^i,a)\right| \label{eq:q4}
\end{align}

Line (\ref{eq:q2}) adds canceling terms and line (\ref{eq:q3}) simplifies the expression using the definition of $\hat{a}^*$.

\vspace{8 pt}

We will now use a concentration inequality to bound (\ref{eq:q4}). To do so we will need refer to some facts about \textit{subgaussian} random variables, which loosely means those whose tails die off at least as fast as a Gaussian.

\newcommand{\subG}[1]{\ensuremath{\textrm{subG}(#1)}}

\textbf{Definition:} $X$ is $\sigma$-subgaussian for some $\sigma > 0$ if $\forall s\in \mathbb{R},~ \E{e^{sX}} \leq e^{\frac{\sigma^2 s^2}{2}}$.

\begin{lemma}[Hoeffding's lemma (1963)]
\label{lemma:1}
Let $X$ be a random variable with mean $0$ and $X\in[a,b]$, $\Delta=b-a$. Then $X$ is $\Delta/2$-subgaussian.
\end{lemma}
The proof is in \cite{ocw_2015_subg}, pg. 20.

\begin{lemma}
\label{lemma:2}
Suppose that $X_1,\ldots ,X_n$ are independent and $\sigma$-subgaussian. Then their mean is $\frac{\sigma}{\sqrt{n}}$-subgaussian.
\end{lemma}
\begin{proof}[Proof of Lemma \ref{lemma:2}]
\begin{align*}
    \E{\exp\left(\frac{s}{n} \sum\limits_{1\le i \le n} X_i\right)} & =
     \E{ \prod\limits_{1\le i \le n}\exp\left(sX_i/n\right) } \\
     &= \prod\limits_{1\le i \le n} \E{\exp\left(sX_i/n\right)} \hspace{1cm} \textrm{by independence} \\
     &= \prod\limits_{1\le i \le n} \exp\left(\frac{\sigma^2s^2}{2n^2}\right) \\
     &= \exp\left( \frac{s^2}{2n^2} \sum\limits_{1\le i \le n}\sigma^2\right) \\
     &= \exp\left( \frac{\left(\sigma / \sqrt{n}\right)^2 s^2}{2}\right)
\end{align*}
\end{proof}



\begin{lemma}
\label{lemma:3}
Suppose that $X_1,\ldots,X_n$ are each $\sigma$-subgaussian. Then 
\begin{equation}
    \E{ \max\limits_{1\leq i \leq n} |X_i| } \leq \sigma \sqrt{2 \log(2n)} 
\end{equation}
\end{lemma}
The proof is in \cite{ocw_2015_subg}, pg. 25.

\vspace{0.5cm}
Now we can apply these results to our setting. $\hat{Q}_\pb^k(\tau^i, a)-Q_\pb(\tau^i,a)$ has mean 0 and bounded width $\Delta$, therefore by Lemma \ref{lemma:1} it is $\Delta/2$-subgaussian. The search procedure performs $N/|\A|$ rollouts per action, so from Lemma \ref{lemma:2} the mean $\hat{Q}_\pb(\tau^i,a) - Q_\pb(\tau^i,a)$ is $\left(\frac{\Delta}{2}\sqrt{\frac{|\A|}{N}} \right)$-subgaussian. Finally, applying Lemma \ref{lemma:3} to the bound in (\ref{eq:q4}), we have

\begin{equation}
    Q_\pb(\tau^i,\hat{a}^*) - V_\pb(\tau^i) \geq -2 \left(\frac{\Delta}{2}\sqrt{\frac{|\A|}{N}} \right) \sqrt{2 \log(2|\A|)}
\end{equation}

We can simplify this a bit using the fact that $\log(X)\leq X$ for $X\geq 1$:

\begin{equation}
\label{eq:t1}
    Q_\pb(\tau^i,\hat{a}^*) - V_\pb(\tau^i) \geq - 2 \Delta |\A|N^{-1/2}
\end{equation}

\vspace{1cm}

We can now prove the main theorem. We denote the policy that follows the MC search procedure up to time $t$ and the blueprint $\pb$ thereafter as $\pi_{s\to t}$. We will prove by induction on $t$  that

\begin{equation}
V_{\pi_{s \to t}}(\tau_0) - V_{\pb} \geq  -2 t\Delta |\A|N^{-1/2}.
\label{eq:ind3}
\end{equation}

The base case is satisfied by definition, since $\pi_{s \to 0} \equiv \pb$.

\vspace{8 pt}

Suppose that at some time $t$, Equation \ref{eq:ind3} is satisfied.

\begin{align}
    V_{\pi_{s \to (t+1)}} & = \mathbb{E}_{\tau_{t+1} \sim \mathcal{P}(\tau_0, \pi_{s \to t})}\left[ \mathbb{E}_{a\sim \pi_s (\tau_{t+1}^i)}\left[ Q_\pi(\tau_{t+1}^i, a)\right | \tau_{t+1}] \right] \\
& = \mathbb{E}_{\tau_{t+1} \sim \mathcal{P}(\tau_0, \pi_{s \to t}), a\sim \pi_s(\tau_{t+1}^i)}\left[ Q_\pb(\tau_{t+1}^i, a) \right] & \textrm{(Law of Total Exp.)}\\
& \geq \mathbb{E}_{\tau_{t+1} \sim \mathcal{P}(\tau_0, \pi_{s \to t})}\left[ V_\pb(\tau_{t+1}^i) \right] -2 \Delta |\A|N^{-1/2} & \textrm{(Eq. \ref{eq:t1})} \\
& = V_{\pi_{s \to t}}(\tau_t^i) -2 \Delta |\A|N^{-1/2} \\
& \geq V_\pb -2 (t+1)\Delta |\A|N^{-1/2} & \textrm{(Eq. \ref{eq:ind3})}
\end{align}

Rearranging,

\begin{equation}
    V_{\pi_{s\to(t+1)}} - V_\pb \geq -2 (t+1)\Delta |\A|N^{-1/2}
\end{equation}

This completes the proof by induction. Since the game length is bounded by $T$, $V_{\pi_{s\to T}} \equiv V_{\pi_s}$ and the proof is complete.






\end{proof}

\twocolumn

\end{document}